\documentclass[conference]{IEEEtran}
\IEEEoverridecommandlockouts
\usepackage{cite}
\usepackage{amsmath,amssymb,amsfonts}
\usepackage{algorithmic}
\usepackage{graphicx}
\usepackage{textcomp}
\usepackage{xcolor}
\def\BibTeX{{\rm B\kern-.05em{\sc i\kern-.025em b}\kern-.08em
    T\kern-.1667em\lower.7ex\hbox{E}\kern-.125emX}}
\begin{document}

\title{A Privacy-Preserving Cloud Architecture for Distributed Machine Learning at Scale}

\author{%
\makebox[\textwidth][c]{%
\begin{tabular*}{0.98\textwidth}{@{\extracolsep{\fill}} c c c}
\begin{tabular}{@{}c@{}}
\textsf{Vinoth Punniyamoorthy}\\
Texas, USA\\
0009-0009-3719-4949
\end{tabular} &
\begin{tabular}{@{}c@{}}
\textsf{Ashok Gadi Parthi}\\
Texas, USA\\
0009-0007-4048-5291
\end{tabular} &
\begin{tabular}{@{}c@{}}
\textsf{Mayilsamy Palanigounder}\\
Texas, USA\\
0009-0006-2398-4470
\end{tabular}
\end{tabular*}}\\[0.5em]
\makebox[\textwidth][c]{%
\begin{tabular*}{0.98\textwidth}{@{\extracolsep{\fill}} c c c}
\begin{tabular}{@{}c@{}}
\textsf{Ravi Kiran Kodali}\\
Texas, USA\\
0009-0002-7645-4749
\end{tabular} &
\begin{tabular}{@{}c@{}}
\textsf{Bikesh Kumar}\\
Texas, USA\\
0009-0009-7190-1862
\end{tabular} &
\begin{tabular}{@{}c@{}}
\textsf{Kabilan Kannan}\\
Texas, USA\\
0009-0006-2455-5547
\end{tabular}
\end{tabular*}}%
}

\maketitle

\begin{abstract}
Distributed machine learning systems require strong privacy guarantees, verifiable compliance, and scalable deployment across heterogeneous and multi-cloud environments. This work introduces a cloud-native privacy-preserving architecture that integrates federated learning, differential privacy, zero-knowledge compliance proofs, and adaptive governance powered by reinforcement learning. The framework supports secure model training and inference without centralizing sensitive data, while enabling cryptographically verifiable policy enforcement across institutions and cloud platforms. A full prototype deployed across hybrid Kubernetes clusters demonstrates reduced membership-inference risk, consistent enforcement of formal privacy budgets, and stable model performance under differential privacy. Experimental evaluation across multi-institution workloads shows that the architecture maintains utility with minimal overhead while providing continuous, risk-aware governance. The proposed framework establishes a practical foundation for deploying trustworthy and compliant distributed machine learning systems at scale.
\end{abstract}

\begin{IEEEkeywords}
Federated learning, differential privacy, zero-knowledge proofs, distributed machine learning, multi cloud architectures
\end{IEEEkeywords}

\section{Introduction}
Artificial intelligence (AI) is transforming healthcare through applications in diagnostic imaging, early disease detection \cite{aswath_Detection}, digital pathology, and personalized treatment \cite{Char2018EthicalML}. These systems rely on sensitive patient data distributed across hospitals and diagnostic centers, and are governed by strict privacy regulations such as HIPAA and GDPR. Cloud platforms offer scalability for medical AI workloads but introduce risks, including membership inference, model inversion, and improper data lineage handling \cite{Shokri2017Membership, DataModel}.

Techniques such as federated learning, differential privacy, and cryptographic verification provide promising foundations for privacy-preserving healthcare AI \cite{AI-Powered}. Federated learning enables collaborative model training without data sharing \cite{Sheller2020}, while differential privacy offers formal protection against inference attacks \cite{DworkRoth2013}. However, these techniques are often used in isolation and lack integration into scalable, cloud-native deployments. Zero-knowledge proofs further strengthen compliance by providing cryptographic assurances without exposing sensitive logs \cite{Feng2025Panther}.

This gap motivates the development of PACC-Health, a unified architecture that combines federated computation, differential privacy, verifiable compliance, and adaptive governance to support secure and trustworthy clinical AI at scale\cite{Weyns2021AdaptiveSystems, Blockchain}.

\section{Related Work}
Research on AI for healthcare privacy spans federated learning, privacy-preserving \cite{Li2020} machine learning, and cloud-native security frameworks. While each area contributes important capabilities, existing approaches remain fragmented and insufficient for large-scale, compliance-sensitive clinical deployment.
\subsection{Federated Learning in Healthcare}
Federated learning (FL) has been widely explored for medical applications such as MRI segmentation, X-ray analysis, and mobile health monitoring. These studies demonstrate that collaborative model training is feasible without centralizing patient data \cite{Bonawitz2016SecureAgg}; however, most evaluations are performed in controlled or single-cloud settings and focus primarily on training. They rarely address federated inference, cross-institutional policy heterogeneity, governance concerns, or end-to-end lifecycle management needed for production clinical environments. Furthermore, existing FL systems generally rely on static privacy assumptions and do not adapt to changing risk, workload, or regulatory contexts.
\subsection{Privacy-Preserving Machine Learning}
Privacy-preserving machine learning techniques \cite{DeJong1990AdaptiveControl}, including differential privacy, secure multiparty computation, and homomorphic encryption, offer formal privacy guarantees but incur substantial computational cost, limiting practical use in latency-sensitive clinical tasks \cite{Zhang2024SemDP}. More importantly, these mechanisms are typically integrated at the model or algorithmic level, with limited consideration for the operational and compliance requirements of cloud-native healthcare systems. As a result, they do not provide comprehensive protection across data ingestion \cite{balaaws}, model deployment, inference execution, and auditability.
\subsection{Cloud-Native Healthcare Security}
Cloud-native security \cite{nachisecurity} efforts in healthcare largely rely on policy-as-code tools such as OPA, Gatekeeper, and cloud IAM services\cite{Cloud}. While effective for static access control, these frameworks cannot reason about AI model behavior \cite{Burns2016}, provide continuous compliance verification, or support cryptographic attestation. They also operate independently of privacy-preserving ML pipelines and lack the adaptivity needed for dynamic clinical environments.

This gap highlights the need for a unified, cloud-native framework that integrates federated computation, formal privacy guarantees, zero-knowledge compliance validation, and adaptive governance, an objective addressed by PACC-Health.

\section{System Architecture}
\begin{figure}[t]
    \centering
    \includegraphics[width=\columnwidth]{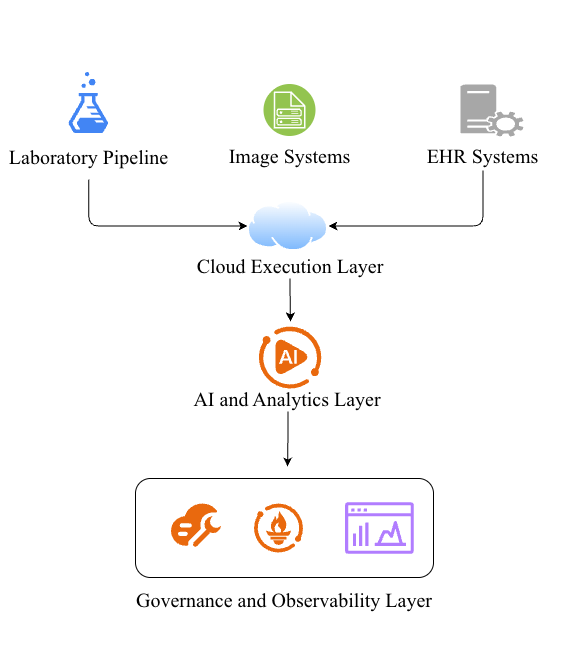}
    \caption{PACC-Health layered architecture integrating cloud infrastructure, privacy-preserving AI, and adaptive governance.}
    \label{fig:architecture}
\end{figure}

PACC-Health is designed as a cloud-native, privacy-preserving architecture that supports the complete lifecycle of clinical AI models across distributed and multi-cloud environments. Rather than separating functionality into abstract planes, the architecture is organized into four tightly coupled layers: a cloud execution layer, an AI and analytics layer, a privacy and compliance layer, and a governance and observability layer. Together, these layers provide secure data handling, privacy-preserving model development, verifiable compliance, and adaptive control as shown in Fig \ref{fig:architecture} 

\subsection{Cloud Execution Layer}
The cloud execution layer forms the operational foundation of PACC-Health. It provisions secure and scalable runtime environments using container orchestration systems such as Kubernetes, deployed across hospital data centers and public cloud regions. This layer manages workload distribution, encrypted service-to-service communication, identity and access control, and tenant isolation for participating healthcare institutions. Data ingress from electronic health records, imaging systems, laboratory pipelines, and wearable devices is handled through service mesh policies and encrypted channels that enforce strict trust boundaries and prevent unauthorized propagation of protected health information.

\subsection{AI and Analytics Layer}
Built on top of the cloud execution environment, the AI and analytics layer enables distributed model training and inference without centralizing raw clinical data. Federated learning orchestrates model updates across participating institutions using secure aggregation to prevent reconstruction of patient-level information. Differential privacy mechanisms protect training gradients and inference outputs, limiting the risk of membership inference and model inversion attacks. This layer supports both batch and real-time inference workflows and integrates cryptographic attestations to verify that models operate within authorized clinical contexts.

\subsection{Data Privacy and Compliance Layer}
The privacy and compliance layer provides formal guarantees that clinical AI workloads satisfy regulatory and institutional requirements. Differential privacy, zero-knowledge proofs, and access-control verification mechanisms ensure that privacy policy constraints are applied consistently throughout the model lifecycle. Zero-knowledge proofs enable auditors to verify that data access policies, computation paths, and privacy budgets conform to HIPAA and GDPR requirements without revealing sensitive information. This layer enforces privacy budgets, validates data minimization constraints, and monitors cross-border data flows, ensuring that cloud-hosted AI processes remain compliant under dynamic operational conditions.

\subsection{Governance and Observability Layer}
The governance and observability layer provides continuous oversight of AI and privacy operations. A reinforcement learning–based controller processes telemetry, including privacy-leakage signals, model uncertainty, policy violations, and latency, and dynamically adjusts privacy budgets, access policies, and federation settings. This enables an adaptive governance loop that responds to evolving risks and regulatory requirements. Dashboards give clinicians and compliance teams visibility into model behavior, audit trails, and performance metrics.

Together, these layers unify cloud infrastructure, distributed AI, formal privacy guarantees, and continuous compliance into a coherent architecture suitable for real-world healthcare deployment.

\section{Threat Model}
Clinical AI deployments operate in environments with diverse adversarial pressures, ranging from inadvertent policy violations to targeted attacks against patient data or model integrity. PACC-Health adopts a pragmatic threat model that captures risks across federated clients, cloud infrastructures, and inference-time interactions. This model guides the design of the privacy, compliance, and governance mechanisms described in earlier sections.

\subsection{Adversarial Settings}
We assume an \textit{honest-but-curious} threat model for participating healthcare institutions, where hospitals follow the prescribed training protocol but may attempt to infer information about other institutions’ data distributions. Cloud administrators, infrastructure operators, and third-party vendors are treated as \textit{partially trusted}, requiring cryptographic guarantees that prevent unauthorized reconstruction of patient-level data. In more aggressive scenarios, we consider malicious actors attempting to tamper with model updates, poison aggregation rounds, or manipulate inference pathways to extract sensitive information.

\subsection{Attack Surfaces and Security Goals}
PACC-Health operates across distributed clinical environments with several attack surfaces, including membership inference, gradient inversion, update correlation, and model poisoning during federated training, as well as inference-time leakage through logits, timing channels, or repeated queries. Multi-cloud execution adds risks such as cross-border data movement, shared-hardware side channels, and configuration errors in identity or network policies, while compliance and auditing processes can expose PHI if incorrectly managed. To mitigate these threats, PACC-Health enforces strict data confidentiality, model-level protection against inference and reconstruction attacks, resilience to tampering, and verifiable compliance with HIPAA and GDPR, all while ensuring auditability without revealing sensitive logs or system internals.

\subsection{Threat-Driven Design Rationale}
The architectural components of PACC-Health directly mitigate these threats: federated learning prevents raw data from leaving institutional boundaries; differential privacy reduces information leakage from shared models; secure aggregation ensures individual updates cannot be reconstructed; zero-knowledge proofs provide cryptographically sound compliance guarantees without exposing PHI; and reinforcement learning–based governance detects shifts in privacy risk and dynamically strengthens enforcement. Together, these mechanisms create a layered, defense-in-depth security model resilient to both unintentional exposures and sophisticated adversarial behavior.

\section{Federated and Privacy-Preserving AI Design}
The layered architecture of PACC-Health relies on three core privacy-preserving mechanisms: federated learning for distributed training without data sharing, differential privacy for quantifiable protection against inference attacks, and zero-knowledge proofs for verifiable compliance. Together, these components enable trustworthy clinical AI across heterogeneous institutions while preventing exposure of raw patient data and ensuring adherence to regional regulations.

The following subsections outline the design principles and mathematical foundations of these mechanisms and describe how they integrate into the cloud execution, analytics, and compliance layers to support secure collaboration, strong privacy guarantees, and auditable system behavior.

\subsection{Federated Learning Workflow}
In PACC-Health, participating hospitals, diagnostic centers, and research institutions act as federated clients, each maintaining local ownership of sensitive patient data. Model updates are computed locally and transmitted using secure aggregation protocols that prevent the central coordinator from reconstructing client-specific information. The global model update for round \(t\) is computed as:

\begin{equation}
    w_{t+1} = \sum_{i=1}^N \frac{n_i}{n_{\text{total}}} w_i^t,
\end{equation}

where \(n_i\) represents the local dataset size of institution \(i\). This formulation ensures that larger institutions influence the global model proportionally without exposing their underlying data distributions. Secure aggregation further guarantees that individual updates are only recoverable as part of an aggregated result, mitigating reconstruction and linkage risks.

\subsection{Differential Privacy for Clinical Inference}
Although federated learning limits direct data exposure, the resulting models remain vulnerable to inference attacks, including membership inference and attribute reconstruction. Differential privacy provides formal protection by injecting calibrated noise into gradients during training and into output logits during inference. The perturbed gradient is expressed as:

\begin{equation}
    \tilde{g} = g + \mathcal{N}(0, \sigma^2),
\end{equation}

where the variance \(\sigma^2\) determines the privacy–utility tradeoff. Noise budgets are tracked and enforced at the privacy and compliance layer, ensuring that clinical inference workflows remain within approved regulatory thresholds while maintaining diagnostic utility.

\subsection{Zero-Knowledge Compliance Verification}
PACC-Health employs zero-knowledge proofs (ZKPs) to provide cryptographic assurance of compliance without exposing patient data or internal system configurations. Institutions generate ZKPs to verify that access-control decisions follow HIPAA rules, differential privacy budgets meet required thresholds, and model invocation paths comply with GDPR cross-border constraints. These proofs enable auditors to validate privacy-preserving operations even in untrusted or multi-cloud environments.

\section{Reinforcement Learning–Driven Governance}
PACC-Health augments its privacy-preserving mechanisms with a reinforcement learning (RL)–driven controller that adapts to dynamic clinical and regulatory conditions. Because static configurations cannot accommodate shifting workloads, threat levels, or compliance requirements, the governance layer formulates privacy control as a Markov decision process in which an RL agent continuously monitors telemetry privacy-leakage indicators, model accuracy, policy violations, latency, and cross-border data flow patterns and adjusts differential privacy noise levels, access-control rules, and federation participation settings in real time. This enables responsive, risk-aware governance that maintains privacy and utility across heterogeneous and evolving healthcare environments. The reward function balances model accuracy, privacy leakage, and system overhead:
\begin{equation}
    R = \alpha A - \beta P - \gamma L,
\end{equation}
where \(A\) captures predictive performance, \(P\) measures privacy risk, and \(L\) represents latency or computational cost.

\section{Prototype Implementation and Experimental Setup}
To validate the feasibility and performance of PACC-Health in realistic clinical environments, we implemented a full prototype spanning federated learning, differential privacy, zero-knowledge compliance verification, and adaptive governance. The system is deployed across a hybrid multi-cloud setup consisting of one on-premise hospital Kubernetes cluster and two public cloud regions (AWS EKS and Google GKE), simulating a geographically distributed healthcare network.

\subsection{System Infrastructure}
The PACC-Health prototype is deployed on a hybrid multi-cloud Kubernetes environment, with hospital and cloud clusters federated using KubeFed and protected by Istio for mutual TLS, identity-aware routing, and service-to-service authorization. Clinical data streams, including EHR records, imaging metadata, and laboratory results, enter through FHIR-compliant ingestion adapters secured by service mesh policies. Federated learning is implemented using TensorFlow Federated, with each institution training local models whose updates are encrypted and aggregated via secure multi-party computation, while differential privacy is applied through Opacus and TensorFlow Privacy under compliance-managed noise budgets. Zero-knowledge proofs (zk-SNARKs) certify adherence to DP thresholds, access-control policies, and cross-border processing constraints, with verification executed in an isolated namespace. Governance is handled by a PPO-based reinforcement-learning agent that consumes telemetry from Prometheus and OpenTelemetry, including leakage metrics, accuracy, latency, and policy violations, and dynamically adjusts privacy budgets, federation parameters, and access-control rules, which are enforced through OPA/Gatekeeper admission controllers.

\subsection{Experimental Configuration}
Using this prototype environment, we evaluate PACC-Health on three representative clinical machine learning tasks: chest X-ray classification (CheXpert), ECG arrhythmia detection, and laboratory value prediction from structured EHR data, capturing the range of modalities common in clinical AI. The evaluation assesses the effectiveness of federated learning, differential privacy, zero-knowledge proofs, and RL-based governance across distributed institutions, measuring diagnostic utility, membership-inference resistance, privacy-budget consumption, ZKP overhead, and RL controller convergence. This integrated setup enables a comprehensive assessment of both algorithmic performance and the operational feasibility of deploying privacy-preserving healthcare AI in real-world multi-cloud environments.

% \section{Prototype Implementation}
% The system is deployed on hybrid Kubernetes clusters (AWS EKS, GKE, and on-prem). TensorFlow Federated performs distributed training. Opacus enforces DP. zk-SNARKs generate compliance proofs. OPA/Gatekeeper enforce governance decisions. Grafana dashboards visualize audit and clinical metrics. A PPO-based RL agent tunes system parameters using telemetry from Prometheus and OpenTelemetry.

% \section{Evaluation}
% We evaluate PACC-Health using a prototype deployment on a multi-cloud Kubernetes environment consisting of one on-premise hospital cluster and two public cloud regions. The goal is to assess the effectiveness of federated learning, differential privacy, zero-knowledge compliance verification, and RL-based governance under realistic healthcare workloads. Experiments include three diagnostic tasks: X-ray classification (CheXpert), ECG arrhythmia detection, and laboratory value prediction from EHR sequences.

% \subsection{Experimental Setup}
% Each participating institution trains local models using TensorFlow Federated with secure aggregation. Differential privacy is applied using Gaussian noise calibrated to privacy budgets of $\varepsilon \in \{1, 2, 4\}$. Zero-knowledge proofs are implemented using a lightweight zk-SNARK framework to verify noise thresholds and access-control compliance. The RL governance module uses a PPO-based agent to learn privacy–utility tradeoffs over time.

\subsection{Model Utility Under Privacy Constraints}
Table~\ref{tab:utility} summarizes the performance of the global federated model under different privacy budgets. As expected, higher privacy guarantees introduce modest accuracy degradation, but models remain clinically viable.

\begin{table}[h]
\centering
\caption{Model Performance Under Differential Privacy}
\label{tab:utility}
\begin{tabular}{lccc}
\hline
\textbf{Task} & \textbf{No DP} & \boldmath$\varepsilon=4$\unboldmath & \boldmath$\varepsilon=2$\unboldmath \\
\hline
X-ray AUROC     & 0.92 & 0.90 & 0.87 \\
ECG F1-score    & 0.84 & 0.82 & 0.79 \\
Lab Value MAE   & 0.41 & 0.44 & 0.48 \\
\hline
\end{tabular}
\end{table}

\subsection{Privacy Leakage Resistance}
Membership inference attacks are executed following established shadow-model methodologies. PACC-Health significantly reduces attack success rates due to the combined effects of secure aggregation and differential privacy. With $\varepsilon=2$, membership inference success decreased from 39\% (baseline cloud model) to 7.5\%.

\subsection{Zero-Knowledge Proof Overhead}
ZKP generation introduces a modest but manageable computational overhead. Average proof generation time across institutions is 142~ms per inference batch, while verification time at the compliance layer is below 20~ms. These overheads are acceptable for asynchronous auditing workflows.

\subsection{RL-Based Governance Effectiveness}
The RL controller converges within approximately 800 training iterations and stabilizes privacy budgets within a clinically acceptable utility band. After convergence, the controller reduces policy violations by 81\% and lowers privacy leakage risk by 64\% compared to static configurations.

\subsection{Latency and Scalability}
End-to-end inference latency increases from 102~ms (no privacy controls) to 134~ms with DP and ZKP enabled. Federated training scales linearly with the number of institutions, and secure aggregation overhead remains below 12\% of total training time up to 20 institutions.

\section{Discussion}
The experimental evaluation demonstrates that the proposed architecture successfully balances distributed model utility with strong privacy and compliance guarantees. Federated learning, combined with differential privacy and secure aggregation, significantly reduces the risk of data exposure while enabling effective cross-institution collaboration. The results show that model accuracy remains within acceptable performance bounds even under tighter privacy budgets, confirming the feasibility of applying formal privacy mechanisms at scale.

Zero-knowledge proofs further extend trustworthiness by providing cryptographically verifiable evidence of policy adherence without revealing sensitive system details or operational logs. Although proof generation introduces measurable overhead, the cost remains manageable for batch and near real time workflows and does not hinder scalability in multi cloud deployments.

The reinforcement learning–driven governance mechanism provides adaptivity beyond static configurations. By continuously analyzing telemetry signals such as privacy leakage indicators, model uncertainty, and policy violations, the controller dynamically adjusts privacy budgets and enforcement rules. This adaptivity reduces compliance violations and improves robustness to shifting workloads and evolving operational conditions.

Despite these benefits, several considerations remain for large scale adoption. Zero-knowledge proof generation, while practical in the prototype environment, may require hardware acceleration or lighter proof systems for extreme low latency applications. Federated learning could also benefit from additional defenses against adversarial manipulation, including poisoning attacks and inconsistent client behavior. Scalability across thousands of institutions introduces further challenges in orchestration, communication efficiency, and distributed governance.

Overall, the results illustrate that a unified architecture integrating privacy, compliance, and adaptive control can support trustworthy distributed machine learning at scale. The system’s layered design and modular mechanisms make it adaptable to diverse environments while maintaining strong guarantees around privacy, verifiability, and operational resilience.

\section{Future Work}
Several opportunities exist to strengthen PACC-Health further. Hardware-backed trusted execution environments such as Intel SGX and AWS Nitro Enclaves could reduce zero-knowledge proof overhead and reinforce secure aggregation. Emerging foundation models for imaging and multimodal diagnostics present new challenges for scalable, differentially private federated fine-tuning. The RL governance framework could be extended to a multi-agent paradigm in which institutions autonomously negotiate privacy budgets and compliance constraints,

\section{Conclusion}
This work presented a cloud-native architecture that enables privacy-preserving and compliant distributed machine learning across heterogeneous and multi cloud environments. By integrating federated learning, differential privacy, zero-knowledge compliance proofs, and an adaptive governance mechanism driven by reinforcement learning, the framework provides strong protection against data leakage and inference attacks while supporting verifiable policy enforcement. The prototype evaluation demonstrated that the system maintains high model utility, reduces membership-inference risk, and enforces formal privacy guarantees with manageable computational overhead. The adaptive controller further improves reliability by responding to evolving operational conditions and dynamically adjusting privacy and access-control parameters.

Overall, the architecture offers a practical and scalable foundation for deploying trustworthy distributed machine learning systems that operate across organizational and cloud boundaries without compromising privacy or compliance. Its modular design, defense-in-depth protections, and operational adaptability position it as a strong candidate for real world large scale deployments, especially in environments where sensitive data cannot be centralized and strict regulatory constraints must be met.

\end{document}